\documentclass{article}

\usepackage{arxiv}

\usepackage[utf8]{inputenc} 
\usepackage[T1]{fontenc}    
\usepackage{hyperref}       
\usepackage{url}            
\usepackage{booktabs}       
\usepackage{amsfonts}       
\usepackage{nicefrac}       
\usepackage{microtype}      
\usepackage{lipsum}
\usepackage{graphicx}
\title{Cooperative observation of targets moving over a planar graph with prediction of positions}

\author{
 José E. B. Maia and Levi P. Figueredo \\
  Universidade Estadual do Ceará – UECE \\ 60714-903 - Fortaleza - Ceará - Brasil \\
  \texttt{jose.maia@uece.br, leviportofigueredo@gmail.com} \\
}

\begin{document}
\maketitle

\begin{abstract}
Consider a team with two types of agents: targets and observers. Observers are aerial UAVs that observe targets moving on land with their movements restricted to the paths that form a planar graph on the surface. Observers have limited range of vision and targets do not avoid observers. The objective is to maximize the integral of the number of targets observed in the observation interval. Taking advantage of the fact that the future positions of targets in the short term are predictable, we show in this article a modified hill climbing algorithm that surpasses its previous versions in this new setting of the CTO problem.
\end{abstract}


\section{Introduction}
Unmanned vehicles (UAVs), whether terrestrial, aquatic or aerial, such as drones, already accumulate a variety in civil applications or military defense and attack. Civil applications include environmental monitoring \cite{freeman2015agricultural}, medical assistance \cite{rosser2018surgical}, transport of goods \cite{menouar2017uav}, electronic surveillance \cite{wang2018operation}, and aerial data surveys using photogrammetry techniques or LIDAR \cite{nex2014uav,zhou2009foreword} sensors. Military applications of UAVs already reported include mission of attack \cite{callam2015drone}, defense against attacks by other UAVs \cite{briese2018vision,huh2015vision,molloy2017detection}, reconnaissance \cite{catano2017information} and border surveillance \cite{paucar2018use}.

UAVs are a type of agents suitable for use as observers. The Cooperative Target Observation (CTO) problem domain is one in which a team of moving surveillance robots, for example, drones, must maintain the observation of another target robot team in motion, in order to maximize the Average Number of Observed Targets ($ANOT$) in the period.

The CTO problem domain has a variety of instances depending on the type of movement of the targets, resource constraints, the interaction between targets and observers, and the stated specific objective. The survey in \cite{khan2018cooperative} presents a classification of problems related to CTO.

In this short paper, a new setting and algorithm for the cooperative targets observation problem is presented. In the configuration faced in this work the targets move on a planar graph. For a concrete example, consider an urban scenario in which $N$ aerial UAVs, each with limited observation radius $R$, must patrol $M > N$ targets moving on land. The movement of the UAVs is free while the movement of targets is restricted to certain paths, such as urban roads. Targets are friends who can, for example, be attacked by enemies. In this scenario, it can be assumed that the positions of the targets and the observers, obtained from GPS, are transmitted to a central command, and that the targets are collaborative, not avoiding the presence of the observers.

In this work an algorithm for this problem is developed which adds knowledge of the domain through heuristics to improve the performance of the basic hill climbing algorithm and another already present in the literature.

The CTO problem has been studied in more recent works as in \cite{aswani2017improving,munnangi2020improving,andrade2018organization,figueredo2018search}. However, these publications do not address the target mobility configuration introduced in the present work.

The work is organized as follows: Section 2 presents the methods used in order to solve the problem, Section 3 shows the results obtained and Section 4 concludes.

\section{Methods}
%
In \cite{parker1999cooperative}, Parker formally defines a more general version of this problem, called CMOMMT, for Cooperative Multi-Robot Observation of Multiple Moving Targets, in which the targets are not collaborative and the environment is partially observable. The objective function to be maximized for the observer team is the Average Number of Observed Targets ($ANOT$) in the simulation period, defined by the expression:
\begin{equation}
    ANOT = \frac{1}{T} \sum_{t=0}^T \sum_{j=1}^M \bigvee_{k=1}^N a_{kj}
    \label{eq1}
\end{equation}
where $A = \{a_{ij}\}$ is an $N\times M$ matrix with the $a_{ij} = 1$ if target $j$ is in the sensor range of the observer $i$ and and 0 when not, and $T$ is the number of time-steps of the simulation. The operator $\bigvee_{k=1}^N$ on a column of the A matrix causes each observed target to be counted only once. Since in this work the number of targets is constant in each simulation, $M=24$, the percentage index, normalized by the number of targets in each execution, was used, $\rho = \frac{ANOT}{M}$.

Luke et al \cite{luke2005tunably} defines the CTO problem as a simplified version of the CMOMMT problem in \cite{parker1999cooperative} in which the targets are collaborative and the environment is fully observable. For this problem he proposes centralized and decentralized algorithms, based on k-means and hill climbing, to calculate the trajectory of the observers.

The present work is based on the definition in Luke et al. and the centralized algorithms proposed there. The k-means and hill climbing versions in \cite{luke2005tunably} are taken as baselines for performance comparison with the proposed variant front of the new environment configuration and target movement.

An intuition about the behavior of solutions to the CTO problem, already present in works \cite{parker1999cooperative,luke2005tunably}, is that if two state-space configurations of observer and target positions result in the same value of the index $\rho$ defined in the equation (\ref{eq1}) then the one with larger the total area covered by the sensors, ie, the one in which the average distance between the observers is higher, is preferable. This notion is used in \cite{parker1999cooperative,luke2005tunably} to construct heuristics to solve the problem. In \cite{luke2005tunably}, Luke et al. use this notion to create a variant of the k-means algorithm and compare its performance against that of a hill climbing (HC). In the present work a variant of this notion is used to add a heuristic step to a hill climbing algorithm and its performance is compared to the k-means and hill climbing described in Luke et al. \cite{luke2005tunably}. This algorithm will be denoted by HC+heuristic or HC+h.

However, although there is an improvement in the HC+h algorithm over K-means and HC, it can be seen that this algorithm does not yet use all the information available in that environment. In fact, the targets' future positions are predictable in the short term as they move over a graph. The HC+heuristic+prediction (HC+hp) algorithm works in the same way as HC+h, however, instead of using the target's current positions to determine the next observer destination, HC+hp uses prediction of the target positions some steps forward to assess the merit of the solutions found.

The difference is that in HC+p, observers run behind to follow targets and also do not anticipate changes in their directions when they reach a vertex of the graph, while in HC+hp, observers anticipate the movements of targets by making predictions.

The HC+h algorithm is described here. The reader is referred to article \cite{luke2005tunably} for a description of the other two algorithms used in performance comparison. One step of HC+h is as follows:
\begin{enumerate}
    \item Input: the current state vector, in the state space of the positions of observers and targets, and its value of $\rho$, denoted $\rho_{cur}$.
    \item Output: the next state vector, updating the destinations of the observers.
    \item Generate 100 feasible random perturbations of the vector of states of the observers with uniform distribution U[-10,10] at the X and Y coordinates. Calculate the index $\rho$, denoted $\rho_{new}$, for each new vector of positions. Adopt the best new generated position evaluated by the criterion $\rho_{new}$, if it exists.
    \item If a new best position was not generated, calculate $\rho_{ob}$, the average distance between observers for the disturbed positions with $\rho_{cur}$ = $\rho_{new}$. Adopt the one with the greatest $\rho_{ob}$.
    \item If a new position with the highest $\rho_{ob}$ was not generated, keep the previous observer path in this update cycle.
\end{enumerate}
Step 3 is the basic hill climbing algorithm. Steps 4 and 5 implement the heuristic based intuition described in the previous paragraphs.

The movement of the observers happens in free space while the movement of targets is restricted to being on a planar graph. In the experiments of this article the targets move is random: when a target arrives at a vertex it makes equally probable choice between neighboring vertices to follow.

\section{Results}
The performance was analyzed by simulation in the MASON \cite{masonabout1} environment. Observers walk in the direction indicated by the last received command. The central command does not need to update the trajectory of observers at each step of time. The observer trajectory update (UR) command rate is a relevant parameter because each update places a communication load on the system. The sensitivity to these parameters also measures the robustness of the system, in latu sensu, to losses of messages and other updating faults.

To generate planar graphs, a two-step procedure was used: first, the vertices were generated in random positions and then a Delaunay triangulation algorithm was applied to construct the edges in order to result in a planar graph. All experiments were performed with graphs of 40 vertices.

Figure 1 shows a simulation snapshot in the MASON environment. It shows a generated random graph, circles representing the range of the observers' sensors, and the points on the edges of the graph are moving targets.
\begin{figure}
    \centering
    \includegraphics[scale=0.4]{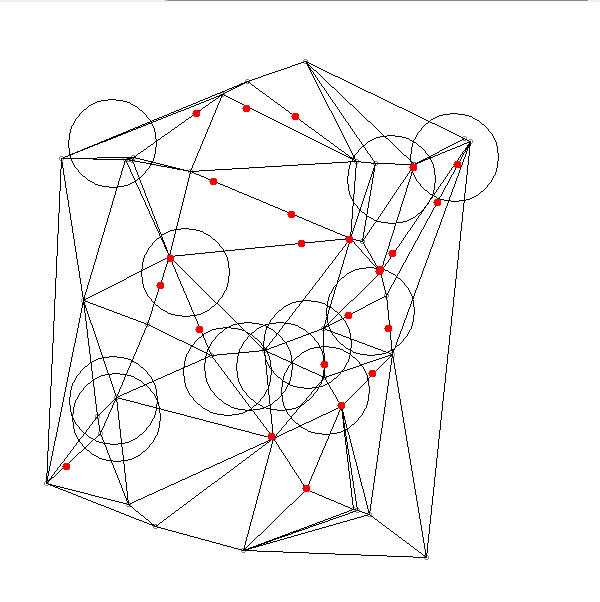}
    \caption{Simulation snapshot in the MASON environment.}
    \label{fig:mason-snapshot}
\end{figure}

In the tests, the simulation used the same parameters adopted in \cite{luke2005tunably}: a rectangular 2D space, with 150x150 units dimension, where targets and observers are inserted; each experiment has a limited time of 1500 time-steps; observers move at 1 unit per time-step, while the targets can move at various speeds RV = {0.1, 0.25, 0.5, 0.75, 0.9} unit per time-step; the sensor range in each observer can be SR = {5, 10, 15, 20, 25} units; and the rate of updating of the trajectory of the observers varies in the set UR = {1, 0.5, 0.25, 0.1, 0.05}; $M=24$ and $N=12$. To collect results, each configuration was simulated 20 runs, each with initial random configuration and independent random number generator seeds. For abbreviation, the algorithms will be referred to as k-means\cite{luke2005tunably}, HC\cite{luke2005tunably} and HC+heuristic or HC+h - for this work. The results are shown in Table 1. 

Table 1 is divided into three parts, where, in each part is varying one of the parameters RV, SR or UR, keeping the others fixed at the median of their set of values. An analysis of this table is as follows.

\begin{table}[ht!]
  \begin{center}
    \caption{The average performance $\rho$ of the HC + heuristic (HC+h) algorithm proposed against K-means + heuristic and Hill Climbing evaluated in \cite{luke2005tunably}, when it varies: sensor range (SR), relative speed between observers and targets (RV), or the update rate of the trajectory of the observers (UR). As the case, UR=0.25, RV=0.50 or SR=15 when the other parameter is varying, N=12 and M=24.}
    \label{tab:table1}
    \begin{tabular}{l|l|l|l|l|l|l}
        \hline\hline
        Sensor range & - & 5 & 10 & 15 & 20 & 25 \\
        \hline
         K-means\cite{luke2005tunably} & m & 0.62 & 0.68 & 0.80 & 0.82 & 0.88 \\
          & sd & 0,043 & 0,048 & 0,036 & 0,034 & 0,027 \\
         HC \cite{luke2005tunably} & m & 0.76 & 0.80 & 0.83 & 0.85 & 0.89  \\
          & sd & 0,066 & 0,058 & 0,054 & 0,061 & 0,060 \\
         HC+h & m & 0.85 & 0.86 & 0.88 & 0.92 & 0.94 \\
          & sd & 0,033 & 0,034 & 0,047 & 0,066 & 0,046 \\
        HC+hp & m & 0.92 & 0.89 & 0.87 & 0.93 & 0.95 \\
          & sd & 0,030 & 0,036 & 0,041 & 0,037 & 0,039 \\
      \hline\hline
       Rel. velocity & - & 0.90 & 0.75 & 0.50 & 0.25 & 0.10 \\
       \hline
         K-means\cite{luke2005tunably} & m & 0.52 & 0.60 & 0.66 & 0.77 & 0.80 \\
          & sd & 0,062 & 0,062 & 0,071 & 0,056 & 0,045 \\
         HC \cite{luke2005tunably} & m & 0.58 & 0.65 & 0.74 & 0.81 & 0.87 \\
          & sd & 0,045 & 0,048 & 0,042 & 0,051& 0,055 \\
         HC+h & m & 0.64 & 0.71 & 0.81 & 0.89 & 0.96 \\
          & sd & 0,028 & 0,046 & 0,033 & 0,052 & 0,048 \\
        HC+hp & m & 0.969 & 0.80 & 0.85 & 0.91 & 0.96 \\
          & sd & 0,029 & 0,039 & 0,032 & 0,039 & 0,038 \\
      \hline\hline
        Update rate & - & 1.00 & 0.50 & 0.25 & 0.10 & 0.05 \\
        \hline
         K-means\cite{luke2005tunably} & m & 0.78 & 0.65 & 0.51 & 0.42 & 0.45 \\
          & sd & 0,043 & 0,044 & 0,057 & 0,026 & 0,046 \\
         HC \cite{luke2005tunably} & m & 0.84 & 0.75 & 0.61 & 0.54 & 0.49 \\
          & sd & 0,054 & 0,058 & 0,052 & 0,029 & 0,049 \\
         HC+h & m & 0.93 & 0.86 & 0.74 & 0.67 & 0.55 \\
          & sd & 0,029 & 0,034 & 0,052 & 0,039 & 0,048 \\
        HC+hp & m & 0.94 & 0.89 & 0.80 & 0.71 & 0.65 \\
          & sd & 0,041 & 0,035 & 0,032 & 0,029 & 0,028 \\
      \hline\hline
    \end{tabular}
  \end{center}
\end{table}
Comparing the rows of the table to the three algorithms, in each comparable region, it is notable that:
\begin{itemize}
        \item The HC+h algorithm presents superior performance for all scenarios.
        \item When the relative velocity (RV) is low or the sensor range (SR) is high or both simultaneously, the difference in algorithm performance is lower. These are less challenging scenarios.
        \item In the high relative velocity (VR) or low range sensor (SR) challenging situations or both simultaneously, it is when the performance of HC+h presents greater resilience while those of the others degrade.
        \item The observer path update rate (UR) strongly affects all algorithms although HC+h maintains performance higher than the others.
        \item The standard deviations between batches of 20 runs did not differ significantly for the algorithms.
\end{itemize}

The total processing time for Table 1 was 93 hours. For real-time execution the maximum duration of one cycle of the HC + h algorithm was 2 s. These numbers are referred to a CORE I7 processor.

\section{Conclusion}
A new configuration of the CTO problem, with the movement of the targets on a planar graph, and two new variants of a centralized algorithm to control the trajectory of the observers are presented and evaluated in this work. A hypothetical motivational example of application was presented in which the edges of the graph are the traffic lanes of an urban region. Performance was analyzed by simulation in MASON \cite{masonabout1}.

Comparative performance table by varying the critical parameters of the problem, ie the range of the sensors (SR), the relative speed between observers and targets (RV), and the update rate of the trajectory of the observers (UR) show that the proposed algorithm variants, HC+h and HC+hp, improved the average performance and slightly reduced the variance against the baseline versions published in \cite{luke2005tunably}. The improvement going from HC+h to HC+hp is the profit of the prediction.

Some work limitations are immediately identified and point to future work. The first is that the proposed algorithm, just like the original, does not construct target motion model. It is believed that incorporating this feature and then applying reinforcement learning to the trajectory of observers will improve the effectiveness of the algorithm. The second is to obtain and execute with real scenario data to counter performance with that obtained in simulation.
\cite{luke2005tunably}

\bibliographystyle{unsrt}  
\bibliography{arxivCTObib}  


\end{document}